\documentclass{svproc}
\usepackage{url}
\usepackage{cite}
\usepackage{amsmath,amssymb,amsfonts}
\usepackage{graphicx}
\usepackage{subfigure}
\usepackage{textcomp}
\usepackage[table,xcdraw]{xcolor}
\usepackage{algorithm}
\usepackage{algpseudocode}
\usepackage{comment}
\usepackage{nomencl}

\usepackage{float}
\usepackage{mathtools}
\usepackage{multirow}
\usepackage{booktabs}

\begin{document}
\mainmatter              % start of a contribution
\pagestyle{empty}
\title{3, 2, 1, Drones Go! A Testbed to Take off UAV Swarm Intelligence for Distributed Sensing}
\titlerunning{A Testbed to Take off UAV Swarm Intelligence for Distributed Sensing}  % abbreviated title (for running head)
\author{Chuhao Qin\inst{1} \and Fethi Candan\inst{2} \and Lyudmila Mihaylova \inst{2} \and Evangelos Pournaras \inst{1}  }
\authorrunning{Chuhao Qin et al.} % abbreviated author list (for running head)
\tocauthor{Chuhao Qin, Fethi Candan, Lyudmila S. Mihaylova and Evangelos Pournaras}
\institute{School of Computing, University of Leeds, Leeds, UK,\\
\email{\{sccq,e.pournaras\}@leeds.ac.uk}
\and
Department of Automatic Control and Systems Engineering, University of Sheffield, Sheffield, UK\\
\email{\{fcandan1,l.s.mihaylova\}@sheffield.ac.uk}
}
\maketitle              % typeset the title of the contribution

\begin{abstract}

This paper introduces a testbed to study distributed sensing problems of Unmanned Aerial Vehicles (UAVs) exhibiting swarm intelligence. Several Smart City applications, such as transport and disaster response, require efficient collection of sensor data by a swarm of intelligent and cooperative UAVs. This often proves to be too complex and costly to study systematically and rigorously without compromising scale, realism and external validity. With the proposed testbed, this paper sets a stepping stone to emulate, within small laboratory spaces, large sensing areas of interest originated from empirical data and simulation models. Over this sensing map, a swarm of low-cost drones can fly allowing the study of a large spectrum of problems such as energy consumption, charging control, navigation and collision avoidance. The applicability of a decentralized multi-agent collective learning algorithm (EPOS) for UAV swarm intelligence along with the assessment of power consumption measurements provide a proof-of-concept and validate the accuracy of the proposed testbed. 

\end{abstract}
\section{Introduction}

Distributed sensing by swarms of drones becomes transformational for Smart Cities. From traffic monitoring, to early detection of forest fires, support of rescue missions under physical disasters and more effective smart farm operations, cooperative drones can deliver at high speed and low cost, high quality data with remarkable resilience to unanticipated environmental and operational conditions~\cite{Floreano2015}. While working with a single high-profile drone is costly and comes with a limited flight range due to power consumption constraints, multiple low-cost cooperative drones distributed over the broader sensing area of interest are a flexible alternative: they can complete a sensing mission in parallel at a shorter period of time during which they are versatile to (re)charge their battery. 

To benefit from this flexibility, drones require coordinated actions with a significant level of autonomy and computational intelligence. Recent decentralized optimization and multi-agent learning algorithms show remarkable scalability and efficiency (low communication and computational cost)~\cite{pournaras2018decentralized, Pournaras2020, jafari2020biologically}, while preserving the privacy and autonomy of software agents. However, designing, prototyping, deploying, testing, and evaluating distributed sensing solutions exhibiting swarm intelligence is a highly complex and interdisciplinary research endeavor. On the one hand, simulation environments reduce the complexity and the number of influential environmental variables, allowing in this way a more targeted study of swarm intelligence algorithms. One the other hand, experimenting with real drones in indoor and even outdoor environments increases realism and external validity~\cite{pournaras2021crowd}. 

To bridge this gap, this paper introduces a testbed to study swarm intelligence of Unmanned Aerial Vehicles (UAVs) for distributed sensing problems. The testbed consists of a 2D area of interest (sensing map). For instance, consider image/video capturing. Such an area can be simply implemented by one large or several smaller interconnected monitors lying on the floor in an indoor lab environment, a projector or another visual medium. These means can project images, videos and more sophisticated visual content of simulation models (e.g. traffic flows~\cite{Gerostathopoulos2019}) to emulate the sensing environment. Low-cost drones can traverse the map and hover to ``scan" points of interest from a certain height. These points form a grid overlay with each cell representing the origin of the collected data (see Fig.~\ref{fig:test-map}). As such, different cells may have different sensing requirements encoding in this way a large spectrum of sensing missions and applications. In this simple and generic context, the proposed testbed allows the study of high Technology Readiness Level (TRL) solutions for a plethora of problems including: charging control of drones, collision avoidance, sensor fusion, autonomous search and navigation, optimization, learning and swarm intelligence algorithms, among others. 

As a proof-of-concept, a multi-agent collective learning approach~\cite{pournaras2018decentralized} is applied to this testbed to coordinate and optimize in a fully decentralized way the navigation and sensing of drones. The optimization cost function minimizes the mismatch between the actual total sensor data collected by drones and the required sensor data, inspired by a real-world scenario of monitoring cycling safety in the city of Zurich~\cite{castells2020cycling}. To validate the realism and applicability of the testbed, the actual energy consumption of the executed navigation and sensing is compared to the estimated energy consumption of the planned navigation and sensing. This estimation is made with a physical model~\cite{stolaroff2018energy}. Results show a significantly low error, demonstrating the capacity of the testbed to move complex swarm intelligence algorithms for UAVs to real-world. 

The contributions of this paper are outlined as follows:
\begin{itemize}
    \item A generic testbed model to study UAV swarm intelligence for distributed sensing.
    \item A first working prototype of the testbed model with a proof-of-concept on accurate estimates of energy consumption in coordinated navigation and sensing. 
    \item A new application domain of multi-agent collective learning~\cite{pournaras2018decentralized} on UAV distributed sensing. 
    \item An open dataset~\cite{Qin2022} collected during experimentation with the testbed. It is generated as a future reference and benchmark to encourage the further development of the testbed by the broader community. 
    \item A rigorous evaluation of the prototyped testbed in an indoor lab environment using real data to demonstrate its applicability and realism.
\end{itemize}

This paper is organized as follows: Section~\ref{sec:related-work} positions this work in literature. Section~\ref{sec:model} introduces the key elements of the proposed testbed model and Section~\ref{sec:prototyping} illustrates a first instantiation of this model. Section~\ref{sec:evaluation} evaluates the proposed testbed and Section~\ref{sec:limitations} discusses its limitation and future extensions. Finally, Section~\ref{sec:conclusion} concludes this paper.

\section{Comparison to Related Work}\label{sec:related-work}

Distributed sensing exhibiting swarm intelligence has been studied in the context of Internet of Things for information sharing and intelligent decision-making~\cite{sun2020survey}. It is used to perform system-wide (global) optimization and solve nonlinear complex problems for the purpose of improving resilience, scalability, flexibility, and privacy (i.e., lower amount but higher quality of data). Swarm intelligence algorithms include particle swarm optimization, flocking, and bio-inspired method.

Swarm intelligence also finds applicability to multi-robot systems to coordinate the actions of large groups of relatively simple robots~\cite{khaldi2015overview}. Recently, UAV systems have become more autonomic and intelligent due to complex missions and dynamic environments. Various swarm intelligence algorithms are applied to the strategic deployment and task execution of a group of autonomous drones,  e.g., traffic monitoring, detection of forests, mapping of physical disasters and others~\cite{zhou2020uav}. The sensing quality and energy constraints are key factors to assess swarm intelligence for distributed sensing.

However, existing swarm intelligence approaches for drones~\cite{fu2019secure, zhu2019multi, chen2018multi} are challenging to implement and assess in real-world. Distributed sensing by a swarm of drones is complex and very expensive to prototype, deploy, test and debug. Licensing is required to operate large drones outdoors. Often promising algorithms are assessed in simulation environments that lack realism and external validity. 

The proposed testbed overcomes this barrier as it can bring together open energy-aware swarm intelligence algorithms, indoor UAVs hardware and abstract sensing maps that can adapt to different application scenarios and context. This comes in sharp contrast to existing testbed solutions\cite{schmittle2018openuav,khan2017mobile} that do not address the energy impact and are limited to an exclusive software-based simulation.  

\section{Testbed Design}\label{sec:model}

This paper introduces a first prototype of an indoor UAV testbed with the aim to support inter-disciplinary research on distributed sensing including: energy-aware distributed learning and optimization algorithms for swarm intelligence, control and communication problems of different UAVs as well as spatio-temporal sensing scenarios for different applications of Smart Cities. The proposed testbed relies on a model, which can be implemented in different lab environments. At an abstract level, the testbed is modeled by the elements presented in the rest of this section. 

\subsection{UAVs}

They communicate to interact with each other directly or via proxies. They run software that implements swarm intelligence for distributed sensing. This software can be customized and configured to serve different sensing applications. Each drone can run its swarm intelligence software within the following continuum~\cite{Fanitabasi2020}: (i) offline/online, remote, centralized computations (server deployment), (ii) offline/online, remote, distributed computations (edge-to-cloud deployment scenario~\cite{Nezami2021}), (iii) online, locally on drones, distributed computations. 

\subsection{Sensing map}

It is the area that UAVs sense at multiple resolutions determined by the height of hovering or flight. At each resolution (height), the map is split into cells, each defining the sensing area of interest. The higher the height, the larger the cell and the lower the sensing quality are likely to be, e.g., image recognition, sound sensitivity, etc.  In the context of a sensing mission, each cell has specific sensing requirements that determine the hovering duration and data acquisition of UAVs. For instance, areas with high traffic may also have higher sensing requirements to accurately capture the traffic flows with drones. In an indoor lab environment, a sensing map can be emulated with one or more monitors or a projector illustrating images, videos and visual outputs of simulations models, e.g. a traffic flow simulation with SUMO~\cite{Gerostathopoulos2019}. In this way, distributed sensing algorithms can be assessed at low-cost, with ease and safety in a wide range of what-if scenarios. 

\subsection{Swarm intelligence}

It plans in a coordinated way the navigation and sensing of multiple UAVs such that the total sensing by the swarm matches well the sensing requirements of all cells. This matching represents the relative approximation between the total sensed values per cell and the actual sensing requirements per cell. Error and correlation metrics such as the root mean squared error, cross-correlation or residuals of summed squares can estimate this matching~\cite{Pournaras2020}. The sensing quality by the swarm is influenced by several factors such as the number of UAVs and their energy consumption as well as the resolution and sensing requirements of the map.

\section{Testbed Prototyping}\label{sec:prototyping}

\begin{figure}[!htb]
\centering
\includegraphics[scale=0.19]{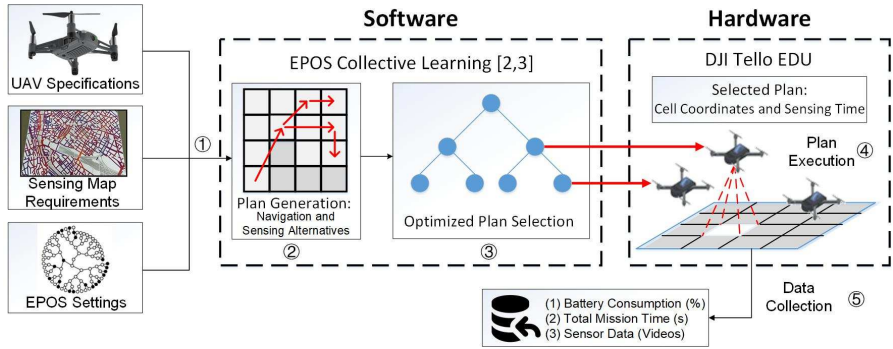}
\caption{An overview of the prototyped testbed architecture.}
\label{fig:main}
\end{figure}

Fig.~\ref{fig:main} illustrates an overview of the testbed architecture. The core of the prototype is the decentralized multi-agent collective learning method of the \emph{Economic Planning and Optimized Selections} (EPOS)~\cite{pournaras2018decentralized,Pournaras2020}. It generates for each agent a finite number of discrete navigation and sensing options, each with an estimated power consumption: the \emph{possible plans} and their \emph{cost} respectively. Plan generation is performed using the UAVs specifications (weight, propeller and battery parameters), grid information about the sensing map and several optimization parameters~\cite{pournaras2018decentralized}. Then the agents interact iteratively in a bottom-up and top-down fashion over a tree communication structure to make a selection such that all choices together add up to maximize the sensing quality (matching). This is shown earlier to be an NP-hard combinatorial optimization problem~\cite{pournaras2018decentralized}. The selected plans contain the X/Y cell coordinates and the sensing (hovering) duration at each cell. They are executed by the UAVs, which record the battery consumption, the total sensing time and the sensor data during the sensing mission.

EPOS is made available as an open-source software~\cite{EPOS}. The collected data are made openly available as a benchmark for the further development of the testbed by the broader community~\cite{Qin2022}. 

\subsection{UAVs and sensing scenario}

\begin{figure}[!htb]
\centering
\subfigure[Drone specification.]{
    \includegraphics[height=2.8cm,width=3.6cm]{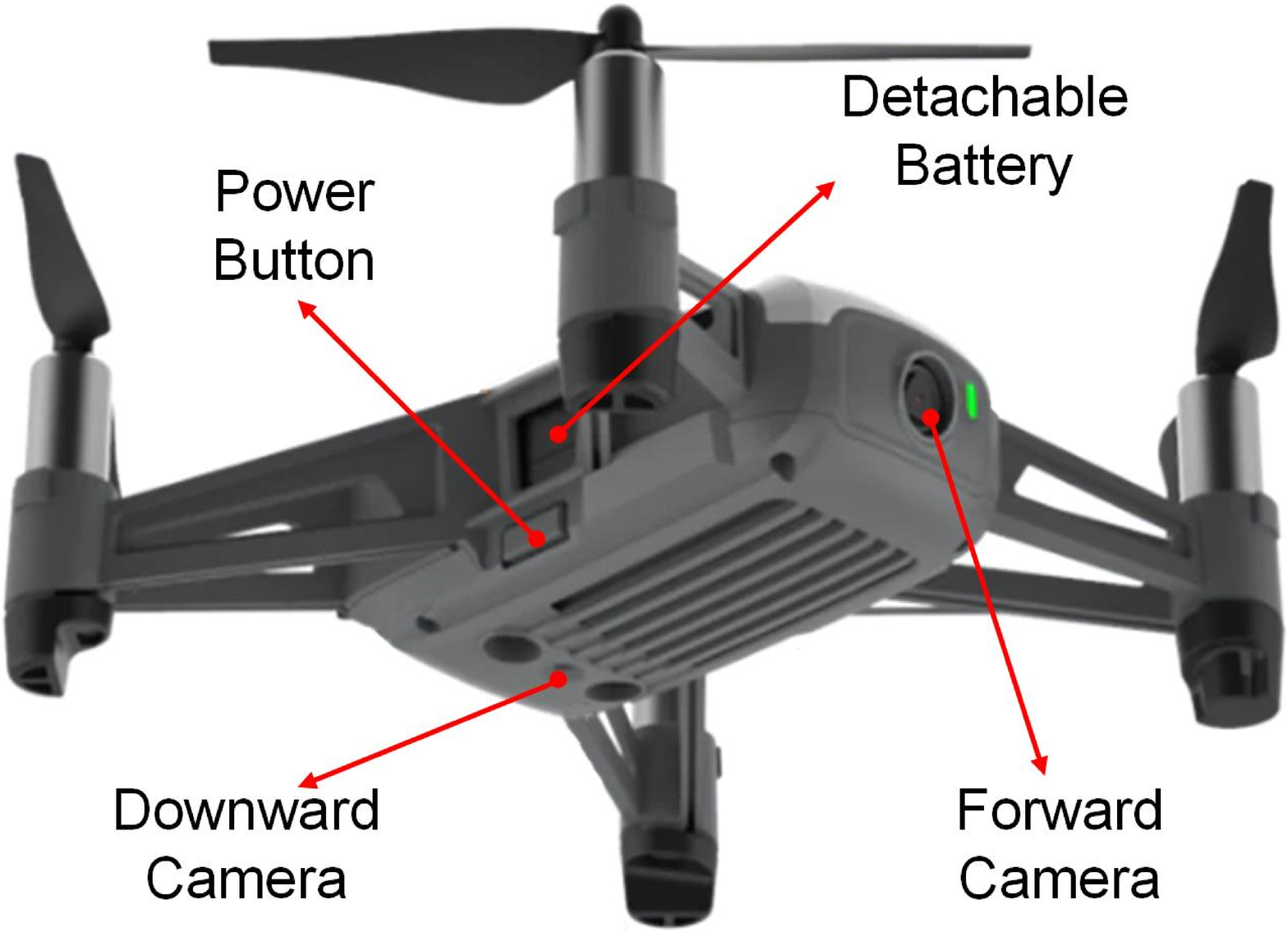}
    \label{fig:Tello_EDU}
}
\subfigure[Camera image cell.]{
    \includegraphics[height=2.8cm,width=3.4cm]{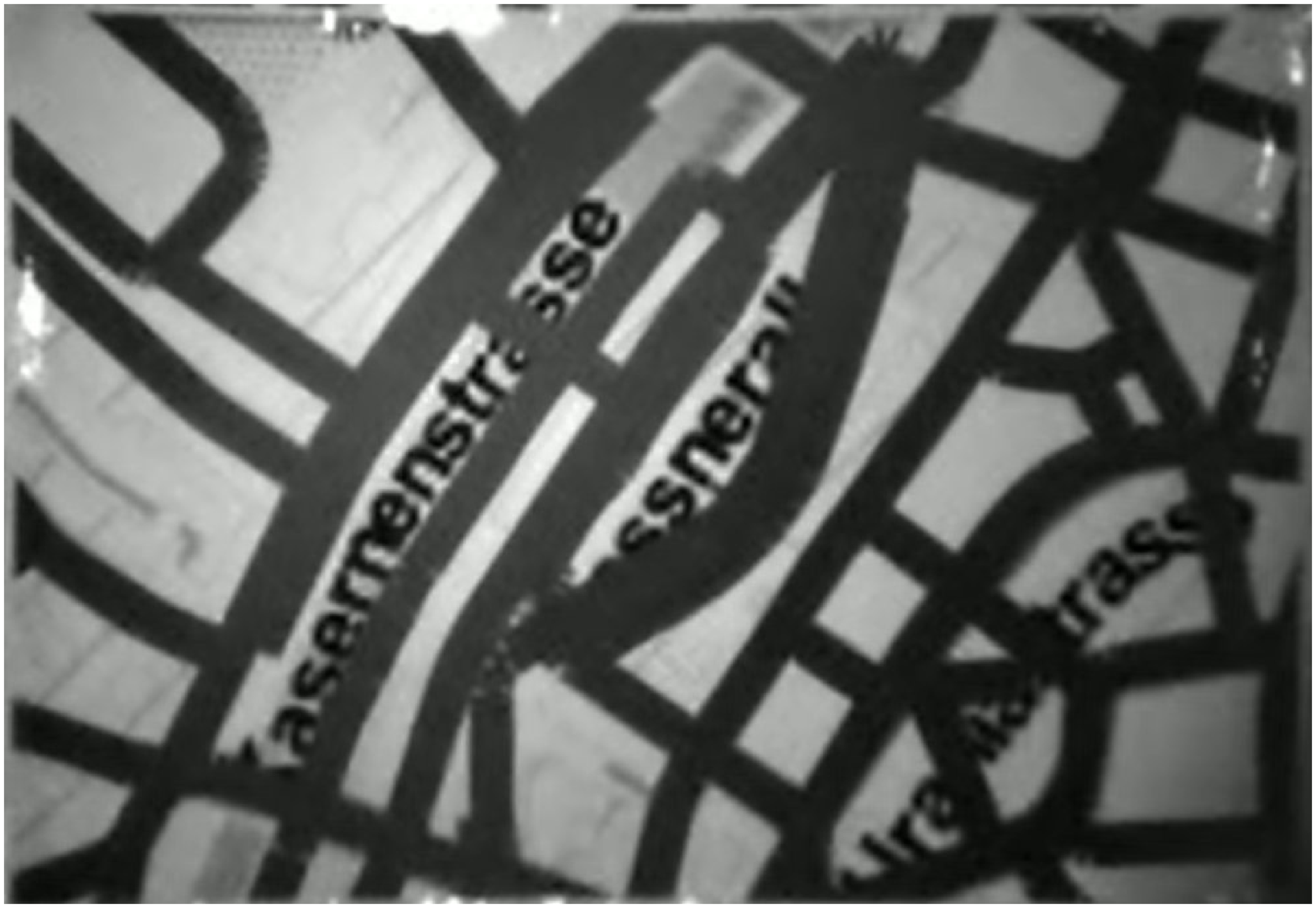}
    \label{fig:cell}
}
\subfigure[Test sensing map.]{
    \includegraphics[height=2.8cm,width=3.7cm]{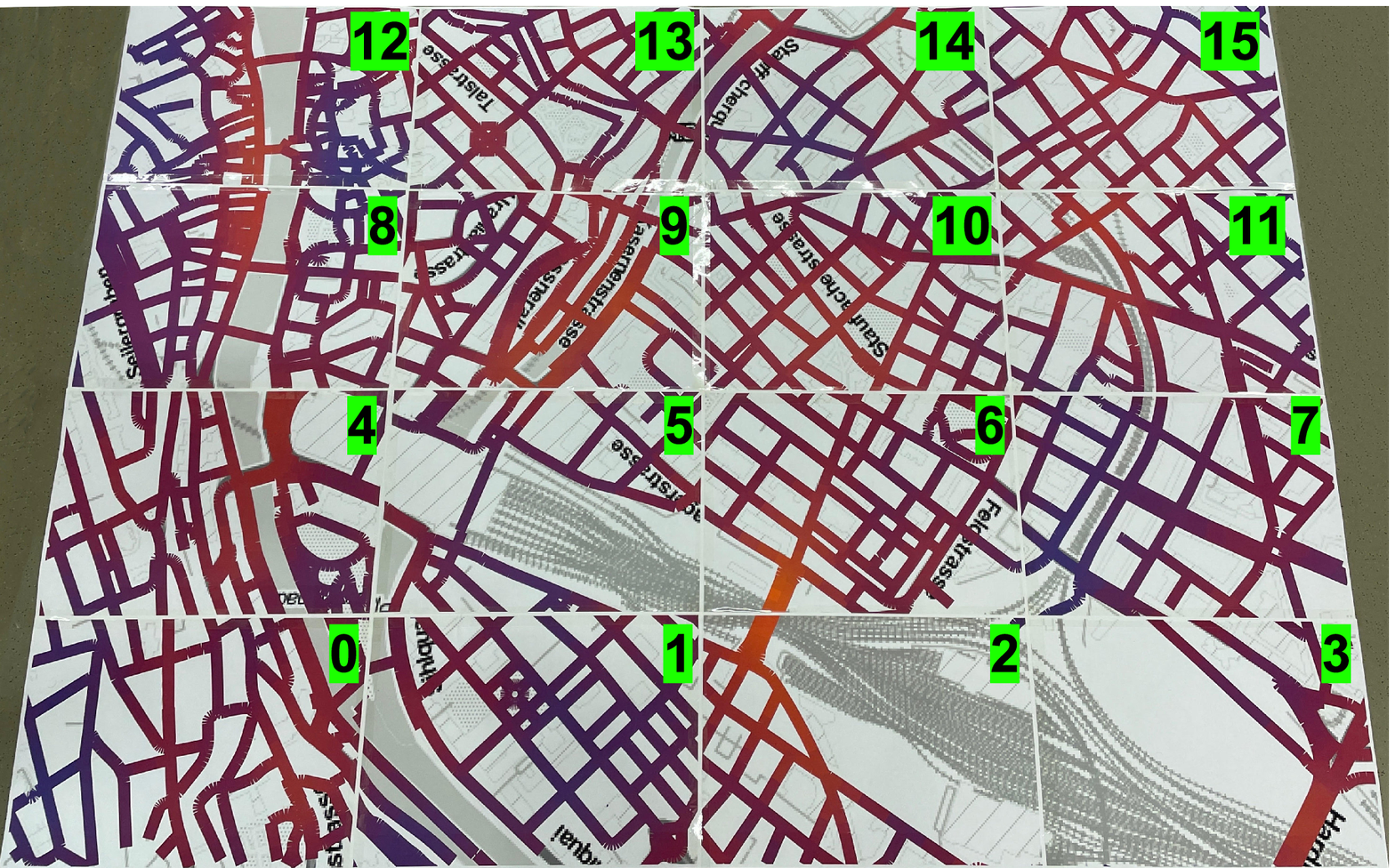}
    \label{fig:test-map}
}
\caption{The DJI Tello EDU drone that flies and hovers to capture images of cells in the sensing map.}
\end{figure}

The DJI Tello EDU UAV is used because of its size, weight, and accessibility~\cite{Tello_UAV, Lipol_Battery, mamchenko2021analysis}. This drone can be programmed in Python, Swift or Scratch, and multiple DJI Tello UAVs can fly for swarm applications. It has $7$ min flight time and precise hovering mode. The drone has 0.1kg weight, 7.26cm propeller length, a battery capacity of $1100 mA$·h and it is configured to fly with an average ground speed of 0.1m/s. Based on this information, both the hovering and maneuvering power can be estimated. DJI Tello UAV has a semi-open-source structure, i.e., the UAV cameras can be accessed by sending commands code, but its internal structure cannot be changed. This drone has 2 different cameras: forward and downward as shown in Fig.~\ref{fig:Tello_EDU}. The forward camera can record RGB $5$ mega-pixel and $1280\times720$ videos, whereas the downward camera, which is used to capture images, can only record gray-scale and $320\times240$ video (see the cell in Fig.~\ref{fig:cell}.

\subsection{Setting up an indoor sensing map}

\begin{figure}[!htb]
\centering
\subfigure[Sensing specification and map.]{
    \includegraphics[scale=0.25]{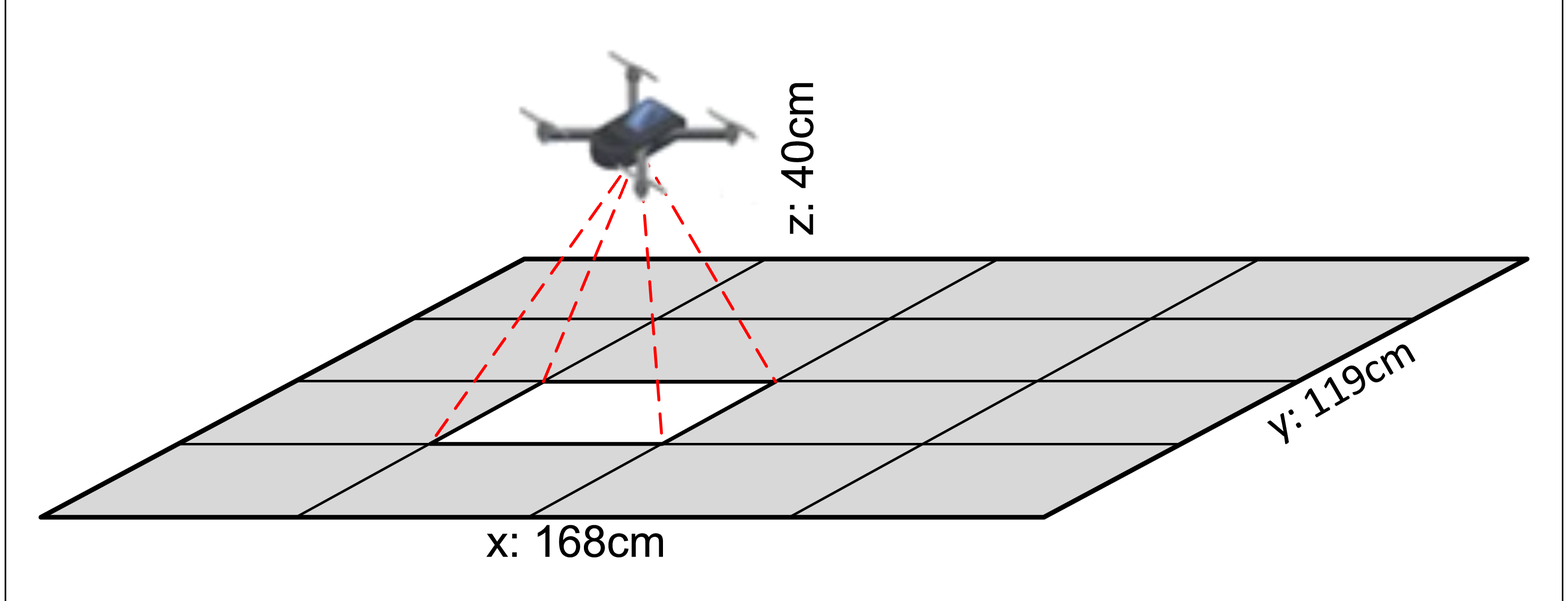}
    \label{fig:cam_pos}
}
\subfigure[Sensing requirements.]{
    \includegraphics[height=2.8cm,width=2.8cm]{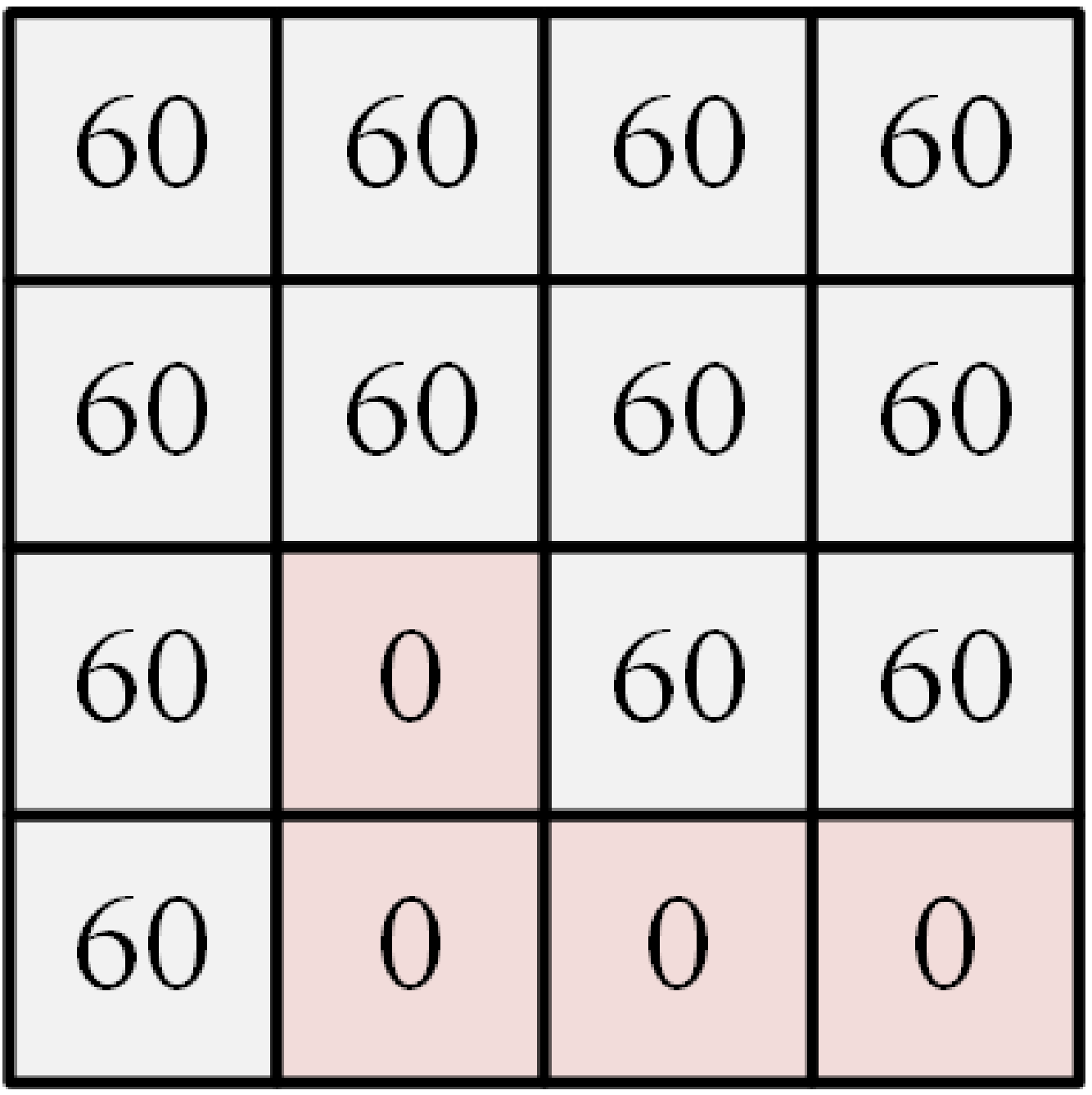}
    \label{fig:sensing_values_map}
}
\caption{Test environment for distributed sensing.}
\label{fig:set}
\end{figure}

A sensing map of size $168 \times 118$ centimeters is tested consisting of $4 \times 4=16$ cells as shown in Fig.~\ref{fig:test-map}. Each cell can be sensed effectively from a height of $40cm$ based on the field of view that the camera of the drone has (see Fig.~\ref{fig:cam_pos}). The map is made by A3 printouts (one per cell) that when put together they show a 2D map of an area in Zurich (see Fig.4 in earlier work~\cite{castells2020cycling}). The cell with index 0 (see Fig.~\ref{fig:test-map}) is the departure/landing cell.

The map is augmented by another layer depicting a continuous kernel density estimation of cycling risk calculated by past bike accident data and other information~\cite{castells2020cycling}. This overlay map can determine the sensing requirements (hovering time) as follows:

\begin{equation}
    t(n) = \frac{f_R(n)}{\sum_{n=0}^{N-1} f_R(n)} \cdot T,
    \label{eq1}
\end{equation}

\noindent where $n$ ($0 \leq n \leq N-1$) denotes the index of a cell on the map (see Fig.~\ref{fig:test-map}) out of a total of $N$ cells, $f_R(n)$ is the kernel density estimate and $T$ is the maximum total operating time of all drones. This means that more data are collected from cells with higher cycling risk, emulating in this way a cycling safety application scenario. For testing, the sensing requirements are simplified by setting each cell with a value of 60 when it contains cycling/road infrastructure, and with 0 when it is does not, see Fig.~\ref{fig:test-map} and Fig.~\ref{fig:sensing_values_map}.

\subsection{Swarm intelligence using the EPOS collective learning}

A number of 10 agents, each one mapping to a drone of the same type, autonomously generate 16 plans that are sequences of 16 real values representing the sensing duration at each cell of the sensing map. The generated plans are random samples of alternative routes with 5 or 6 sensing points that respect battery constraints. The data sampling duration while hovering over a cell is fixed to 13 sec to provide a wide range of alternative routing paths. The plans come with a cost calculated by the total power consumption of flying and hovering~\cite{stolaroff2018energy}. 

The plan selection is made in a coordinated way using EPOS~\cite{pournaras2018decentralized}. For this, agents connect into a balanced binary tree topology within which they interact with their children and parent to improve iteratively their plan selection. The goal of the agents is to minimize the residual of sum squares (RSS) between the following unit-length scaled signals: sensing values per cell summed up over all agents and the sensing requirements per cell. Against this ($\beta=0$), the agents can make a choice that minimizes the power consumption of their drone (baseline, $\beta=1$). Both scenarios are shown to demonstrate the effect of optimizing the sensing quality. The agents perform 40 bottom-up and top-down learning iterations. More information about the collective learning algorithm of EPOS is out of the scope of this paper and can be found in earlier work~\cite{pournaras2018decentralized, Pournaras2020}. 

Table~\ref{tab:epos} summarizes the optimization parameters of EPOS. For this first testbed prototype, the optimization process is performed offline and remotely, however deployments of EPOS for online optimization are already available for future extensions~\cite{Fanitabasi2020}. 

\begin{table}[!htb]
    \centering
    \caption{The test parameters for the optimization of distributed sensing.}
    \label{tab:epos}
        \begin{tabular}{ll}
         \toprule
         \textbf{Parameter} & \textbf{Value} \\
         \midrule
         Number of agents/drones    &   10 \\
         Network topology           &   balanced binary tree \\
         Number of plans per agent  &   16 \\
         Size of possible plans     &   16 \\
         Number of repetitions      &   40 \\
         Preference of agent        &   $\beta = 0$ \\
         Non-linear cost function   &   RSS (unit-length) \\
         \bottomrule
        \end{tabular}
\end{table}

\section{Experimental Evaluation}\label{sec:evaluation}

This section illustrates the measurements performed for the sensing quality as well as the accuracy of energy consumption estimation made during the planning phase. 

Table~\ref{tab:result} shows the measurements made for the optimized navigation and sensing of each drone. The visited cells (see Fig.\ref{fig:sensing_values_map}) are extracted from the selected plans (non-zero values). The actual power consumption of the mission is influenced by the start and end battery level as voltage varies with the LiPo battery capacity~\cite{Lipol_Battery, chuangfeng2011measurement}. The hovering and maneuvering power consumption are calculated based on the earlier power consumption model~\cite{stolaroff2018energy} and that is why it remains constant across the drones with the same specification. 

\newcommand{\tabincell}[2]{\begin{tabular}{@{}#1@{}}#2\end{tabular}}
\begin{table}[!htb]
\centering
\caption{Results of optimizing distributed sensing for each drone: Battery consumption level, visited cells extracted from the selected plan, total traveling time (second), actual power consumption as well as the estimated hovering and maneuvering power (watt).}
\label{tab:result}
\resizebox{\textwidth}{!}{%
\begin{tabular}{lclllcllllllcllll} 
% cccccccccccccc
\toprule
\multirow{2}*{\tabincell{l}{\textbf{UAV} \\ \textbf{Index}}}       & \phantom{abc}     & \multicolumn{3}{l}{\tabincell{l}{\textbf{Battery} \\ \textbf{Level $(\%)$}}} & \phantom{abc}  & \multicolumn{6}{l}{\textbf{Visited Cells Indices}}  & \phantom{abc}     & \multirow{2}*{\tabincell{l}{\textbf{Total} \\ \textbf{Time $(s)$}}}     & \multirow{2}*{\tabincell{l}{\textbf{Actual} \\ \textbf{Power $(w)$}}}     &   \multirow{2}*{\tabincell{l}{\textbf{Hovering} \\ \textbf{Power $(w)$}}}      &   \multirow{2}*{\tabincell{l}{\textbf{Maneuvering} \\ \textbf{Power $(w)$}}}
\\
\cmidrule{3-5} \cmidrule{7-12}
&   & Start & End & Diff. &   & 1st & 2nd & 3rd & 4th & 5th & 6th &   & & & & \\
\midrule
1                       &      & 75  & 47    & 28  &     & 0     &7    & 10     & 12     & 14    & 15   &      & 143.35    & 30.80   & 31.80      & 31.92                      \\
2                        &     & 86 & 69 & 17    &       & 0    & 4    & 6     & 7     & 11    & 12   &        & 135.37    & 30.98   & 31.80      & 31.92                       \\
3                       &      & 99 & 62 & 37      &     & 7    & 9    & 10     & 12     & 14    & 15  &       & 159.32    & 31.42   & 31.80      & 31.92                      \\
4                       &      & 100 & 77 & 23      &      & 4    & 9    & 11     & 12     & 13    & -   &     & 135.31    & 31.59   & 31.80      & 31.92                       \\
5                       &      & 80 & 55 & 25      &      & 4    & 6    & 7     & 9     & 10    & 12   &       & 150.70    & 30.36            & 31.80      & 31.92             \\
6                       &      & 100 & 68 & 32     &       & 4    & 6    & 7     & 8     & 9    & 11   &       & 152.95    & 31.50            & 31.80      & 31.92                         \\
7                       &      & 88  & 75 & 13      &      & 0    & 4    & 8     & 13     & 14    & 15   &      & 130.91    & 30.89           & 31.80      & 31.92                        \\
8                       &      & 100 & 76 & 24      &      & 0    & 8    & 13     & 14     & 15    & -   &      & 125.50    & 31.59           & 31.80      & 31.92                        \\
9                       &      & 100 & 74 & 26      &      & 0    & 6    & 8     & 10     & 11    & 13   &      & 136.22    & 31.59           & 31.80      & 31.92                        \\
10                      &      & 100 & 63 & 37       &     & 8    & 9    & 10     & 11     & 14    & 15  &     & 153.29     & 31.42            & 31.80      & 31.92                        \\ 
\bottomrule
\end{tabular}%
}
\end{table}

\begin{figure}[!htb]
\centering
\subfigure[Coordinated optimized sensing with EPOS that minimizes sensing mismatch.]{
    \includegraphics[scale=0.33]{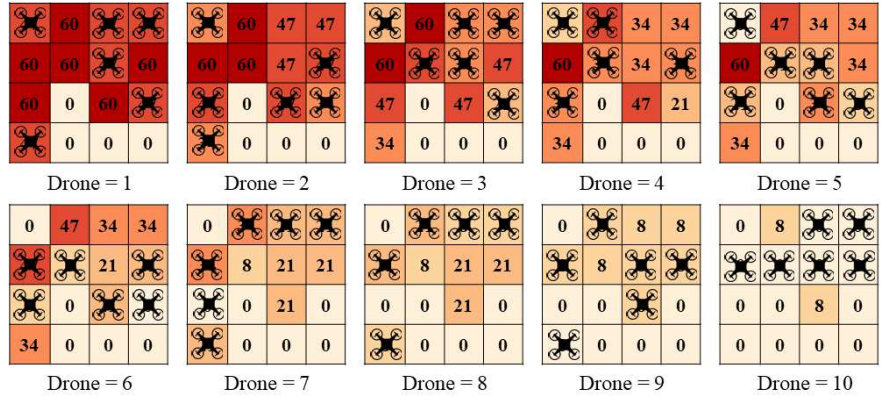}
    \label{fig:sense_flow_mission}
}
\subfigure[Greedy sensing strategy that minimizes energy consumption.]{
    \includegraphics[scale=0.33]{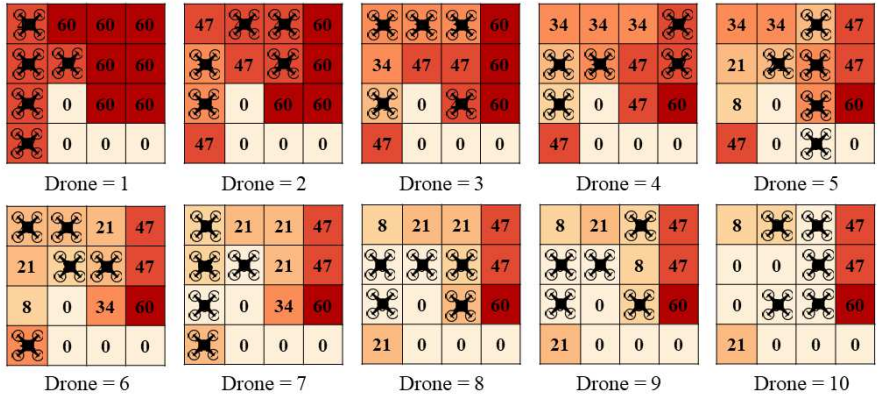}
    \label{fig:sense_flow_energy}
}
\caption{Coordinated vs. greedy sensing by a swarm of drones. The numbers over the series of sensing maps show how the drones meet the target sensing requirements.}
\label{fig:sense_flow}
\end{figure}

Fig.~\ref{fig:sense_flow_mission} illustrates a series of sensing maps depicting how drones meet the target sensing requirements after each trip by minimizing the mismatch between the total actual collected data and the required ones. This strategy ($\beta=0$) is the one implemented in the indoor lab environment. To clearly show the effect of optimized sensing, a greedy strategy ($\beta=1$) is shown in Fig.~\ref{fig:sense_flow_energy} in which drones select the plan with the lowest energy consumption without any coordination. With coordination, the total energy consumption is $35.53kJ$ with a mismatch (RSS) of $0.0057$ and a mission inefficiency of $2.22\%$. The latter is defined by the ratio of required values in all cells that are not sensed by the drones during their mission over the total required values in all cells. In contrast, without coordination, the greedy strategy lowers energy consumption to $27.61kJ$ at a cost of higher mismatch ($0.265$) and mission inefficiency ($26.11\%$). As can be observed in the series of sensing maps, coordination results in routing paths that expand further away from the departure/landing cell, while avoiding oversensing/undersensing that is likely to happen without coordination. 

\begin{figure}[!htb]
\centering
\includegraphics[scale=0.5]{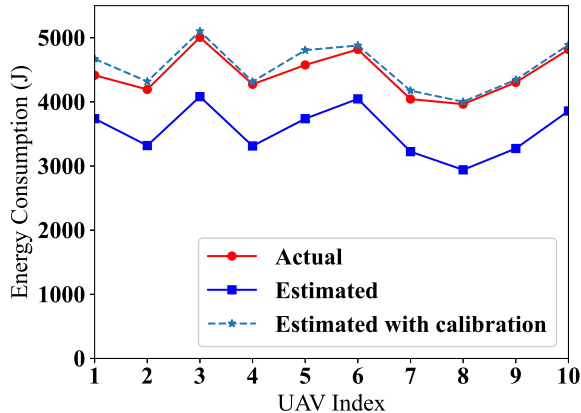}
\caption{Comparison between the estimated and actual energy consumption, with and without calibration.}
\label{fig:energy_results}
\end{figure}

Fig.~\ref{fig:energy_results} compares, for each drone, the actual energy consumption with the model-based estimated one calculated during the planning phase. The actual energy consumption is initially found higher than the estimated one. This is because the drones spend some additional flying time to calibrate between departure and landing. To account for this additional energy consumption, the calibration time is recorded and added up to the original estimated energy consumption, resulting is a highly accurate estimation for all drones. A low error range from 37.41 to 255.52 Joule is attributed to the variable power consumption of the battery. 

\section{Discussion and Limitations}\label{sec:limitations}

The experimental results collected from this first testbed prototype demonstrate its significant applicability and realism to study UAV swarm intelligence for distributed sensing. This is shown by the high sensing quality and accuracy in energy consumption estimation as well as by the testbed flexibility to record complex measurements in an indoor lab environment that emulate several complex outdoor sensing scenarios.

Nevertheless, the testbed can be further improved towards several exciting research avenues: (i) Use of advanced hardware and multimedia to emulate several and more complex sensing maps originated from empirical data and simulation models. (ii) Use of different UAVs with different sensors and data collection capabilities. (iii) Integration of (wireless) charging stations to the testbed to allow the study of longer and more complex sensing missions. (iv) Use of collision avoidance to allow multiple drones to fly in a limited lab space. (v) Use of other swarm intelligence algorithms as well as their integration in online testbed operations. 

\section{Conclusion and Future Work}\label{sec:conclusion}

In conclusion, this paper illustrates a new testbed to study a broad spectrum of research problems on distributed sensing using UAV swarm intelligence. The proposed testbed, setup for indoor lab environments with low-cost drones, simplifies experimentation with complex distributed sensing scenarios that would otherwise require expensive drone equipment, licensing and more uncertainties to handle in unknown outdoor environments. As a proof-of-concept, this paper demonstrates the applicability of a fully decentralized multi-agent collective learning algorithm~\cite{pournaras2018decentralized} to coordinate the navigation and sensing of drones in an emulated scenario of monitoring cycling safety in the city of Zurich. Using the testbed, the accuracy of estimating the energy consumption of drones during navigation and sensing using a physical model is verified. These results show the strong potential of this testbed and opportunities for the broader community to `take off' UAV swarm intelligence research in impactful distributed sensing scenarios of Smart Cities. 

\subsubsection*{Acknowledgements}

This research is supported by the White Rose Collaboration Fund: \emph{Socially Responsible AI for Distributed Autonomous Systems}. Evangelos Pournaras is also supported by a UKRI Future Leaders Fellowship (MR\-/W009560\-/1): \emph{Digitally Assisted Collective Governance of Smart City Commons--ARTIO}. 

\bibliographystyle{unsrt}
\bibliography{EPOS_DRONES}

\begin{thebibliography}{10}

\bibitem{Floreano2015}
Dario Floreano and Robert~J Wood.
\newblock Science, technology and the future of small autonomous drones.
\newblock {\em Nature}, 521(7553):460--466, 2015.

\bibitem{pournaras2018decentralized}
Evangelos Pournaras, Peter Pilgerstorfer, and Thomas Asikis.
\newblock Decentralized collective learning for self-managed sharing economies.
\newblock {\em ACM Transactions on Autonomous and Adaptive Systems (TAAS)},
  13(2):1--33, 2018.

\bibitem{Pournaras2020}
Evangelos Pournaras.
\newblock Collective learning: A 10-year odyssey to human-centered distributed
  intelligence.
\newblock In {\em 2020 IEEE International Conference on Autonomic Computing and
  Self-Organizing Systems (ACSOS)}, pages 205--214. IEEE, 2020.

\bibitem{jafari2020biologically}
Mohammad Jafari, Hao Xu, and Luis Rodolfo~Garcia Carrillo.
\newblock A biologically-inspired reinforcement learning based intelligent
  distributed flocking control for multi-agent systems in presence of uncertain
  system and dynamic environment.
\newblock {\em IFAC Journal of Systems and Control}, 13:100096, 2020.

\bibitem{pournaras2021crowd}
Evangelos Pournaras, Atif~Nabi Ghulam, Renato Kunz, and Regula H{\"a}nggli.
\newblock Crowd sensing and living lab outdoor experimentation made easy.
\newblock {\em IEEE Pervasive Computing}, 21(1):18--27, 2021.

\bibitem{Gerostathopoulos2019}
Ilias Gerostathopoulos and Evangelos Pournaras.
\newblock Trapped in traffic? a self-adaptive framework for decentralized
  traffic optimization.
\newblock In {\em 2019 IEEE/ACM 14th International Symposium on Software
  Engineering for Adaptive and Self-Managing Systems (SEAMS)}, pages 32--38.
  IEEE, 2019.

\bibitem{castells2020cycling}
David Castells-Graells, Christopher Salahub, and Evangelos Pournaras.
\newblock On cycling risk and discomfort: urban safety mapping and bike route
  recommendations.
\newblock {\em Computing}, 102(5):1259--1274, 2020.

\bibitem{stolaroff2018energy}
Joshuah~K Stolaroff, Constantine Samaras, Emma~R O’Neill, Alia Lubers,
  Alexandra~S Mitchell, and Daniel Ceperley.
\newblock Energy use and life cycle greenhouse gas emissions of drones for
  commercial package delivery.
\newblock {\em Nature Communications}, 9(1):1--13, 2018.

\bibitem{Qin2022}
Chuhao Qin, Fethi Candan, Lyudmila Mihaylova, and Evangelos Pournaras.
\newblock {EPOS Optimization datasets for drones}.
\newblock 6 2022.
\newblock doi = {https://doi.org/10.6084/m9.figshare.20069366.v5}.

\bibitem{sun2020survey}
Weifeng Sun, Min Tang, Lijun Zhang, Zhiqiang Huo, and Lei Shu.
\newblock A survey of using swarm intelligence algorithms in {IoT}.
\newblock {\em Sensors}, 20(5):1420, 2020.

\bibitem{khaldi2015overview}
Belkacem Khaldi and Foudil Cherif.
\newblock An overview of swarm robotics: Swarm intelligence applied to
  multi-robotics.
\newblock {\em International Journal of Computer Applications}, 126(2), 2015.

\bibitem{zhou2020uav}
Yongkun Zhou, Bin Rao, and Wei Wang.
\newblock {UAV} swarm intelligence: Recent advances and future trends.
\newblock {\em IEEE Access}, 8:183856--183878, 2020.

\bibitem{fu2019secure}
Zhangjie Fu, Yuanhang Mao, Daojing He, Jingnan Yu, and Guowu Xie.
\newblock Secure multi-{UAV} collaborative task allocation.
\newblock {\em IEEE Access}, 7:35579--35587, 2019.

\bibitem{zhu2019multi}
Moning Zhu, Xiaoxia Du, Xuehua Zhang, He~Luo, and Guoqiang Wang.
\newblock Multi-{UAV} rapid-assessment task-assignment problem in a
  post-earthquake scenario.
\newblock {\em IEEE Access}, 7:74542--74557, 2019.

\bibitem{chen2018multi}
Yongbo Chen, Di~Yang, and Jianqiao Yu.
\newblock Multi-{UAV} task assignment with parameter and time-sensitive
  uncertainties using modified two-part wolf pack search algorithm.
\newblock {\em IEEE Transactions on Aerospace and Electronic Systems},
  54(6):2853--2872, 2018.

\bibitem{schmittle2018openuav}
Matt Schmittle, Anna Lukina, Lukas Vacek, Jnaneshwar Das, Christopher~P
  Buskirk, Stephen Rees, Janos Sztipanovits, Radu Grosu, and Vijay Kumar.
\newblock Open{UAV}: A {UAV} testbed for the cps and robotics community.
\newblock In {\em Proc. of the 2018 ACM/IEEE 9th International Conference on
  Cyber-Physical Systems (ICCPS)}, pages 130--139. IEEE, 2018.

\bibitem{khan2017mobile}
Mouhyemen Khan, Karel Heurtefeux, Amr Mohamed, Khaled~A Harras, and
  Mohammad~Mehedi Hassan.
\newblock Mobile target coverage and tracking on drone-be-gone {UAV}
  cyber-physical testbed.
\newblock {\em IEEE Systems Journal}, 12(4):3485--3496, 2017.

\bibitem{Fanitabasi2020}
Farzam Fanitabasi, Edward Gaere, and Evangelos Pournaras.
\newblock A self-integration testbed for decentralized socio-technical systems.
\newblock {\em Future Generation Computer Systems}, 113:541--555, 2020.

\bibitem{Nezami2021}
Zeinab Nezami, Kamran Zamanifar, Karim Djemame, and Evangelos Pournaras.
\newblock Decentralized edge-to-cloud load balancing: Service placement for the
  internet of things.
\newblock {\em IEEE Access}, 9:64983--65000, 2021.

\bibitem{EPOS}
Evangelos Pournaras.
\newblock {Economic Planning and Optimized Selections (EPOS)}.
\newblock \url{https://github.com/epournaras/epos}, 2022.
\newblock Accessed: 2022-06-27.

\bibitem{Tello_UAV}
Company {DJI Tello}.
\newblock {DJI} {Tello} {EDU}, {Ryzerobotics}.
\newblock \url{https://www.ryzerobotics.com/tello-edu}.
\newblock Accessed: 2022-06-16.

\bibitem{Lipol_Battery}
Company {Fullymax}.
\newblock Fullymax {Battery}.
\newblock \url{http://en.fullymax.com/}.
\newblock Accessed: 2022-06-26.

\bibitem{mamchenko2021analysis}
Mark~V Mamchenko.
\newblock Analysis of control channel cybersecurity of the consumer-grade {UAV}
  by the example of {DJI} {Tello}.
\newblock In {\em Journal of Physics: Conference Series}, volume 1864, page
  012127. IOP Publishing, 2021.

\bibitem{chuangfeng2011measurement}
Huai Chuangfeng, Liu Pingan, and Jia Xueyan.
\newblock Measurement and analysis for lithium battery of high-rate discharge
  performance.
\newblock {\em Procedia Engineering}, 15:2619--2623, 2011.

\end{thebibliography}
\end{document}